\documentclass[11pt,a4paper]{article}
\usepackage[hyperref]{naaclhlt2018}
\usepackage{times}
\usepackage{latexsym}
\usepackage{amsmath}
\usepackage{graphicx}
\usepackage{url}
\usepackage{algorithm}
\usepackage{algpseudocode}
\usepackage{mathrsfs}

\aclfinalcopy % Uncomment this line for the final submission
\def\aclpaperid{1027} %  Enter the acl Paper ID here

\title{Leveraging Intra-User and Inter-User Representation Learning for Automated Hate Speech Detection}

\author{Jing Qian, Mai ElSherief, Elizabeth M. Belding, William Yang Wang\\
  Department of Computer Science\\
  University of California, Santa Barbara\\
  Santa Barbara, CA 93106 USA\\
  {\tt \{jing\_qian, mayelsherif, ebelding, william\}@cs.ucsb.edu} \\}

\date{}

\begin{document}
\maketitle
\begin{abstract}
Hate speech detection is a critical, yet challenging problem in Natural Language Processing (NLP). Despite the existence of numerous studies dedicated to the development of NLP hate speech detection approaches, the 
%\elizabeth{by performance, do you really mean "accuracy"?  I don't think performance is the right word here as it refers to, for instance, time or cost.} 
accuracy is still poor. 
The central problem is that social media posts are short and noisy, and most existing hate speech detection solutions take each post as an isolated input instance, which is likely to yield high false positive and negative rates.
%\elizabeth{Instead of the following clause, how about "which is likely to yield high false positive and negative rates"} 
%which might not be the most ideal setting.  
In this paper, we radically improve 
%\elizabeth{ I'm assuming here that we will demonstrate the radical improvement quantitatively}
%Jing: yes, in the last sentence in the abstract.
automated hate speech detection by presenting a novel model that leverages intra-user and inter-user representation learning for robust hate speech detection on Twitter. In addition to the target Tweet, we collect and analyze the user's historical posts to model intra-user Tweet representations. To suppress the noise in a single Tweet, we also model the similar Tweets posted by all other users with reinforced inter-user representation learning techniques. Experimentally, we show that leveraging these two representations can significantly improve the f-score of a strong bidirectional LSTM baseline model by 10.1\%.
\end{abstract}
\section{Introduction}
The rapid rise in user-generated web content has  not only yielded a vast increase in information accessibility, but has also given individuals an easy platform on which to share their beliefs and to publicly communicate with others.  Unfortunately, this has also led to nefarious uses of online spaces, for instance for the propagation of hate speech.  
%Due to the negative emotional impacts of hate speech, hate speech detection, in part for its timely removal from social media sites, has become a critical task.
%In recent years, with the rapidly rising volume of the Internet, user-generated web content has been massively increasing. However, this also leads to the growing amount of hate speech on the social media. Since a hate speech post may cause negative effects on the others and our society, hate speech detection becomes a task of significant social impact.  
\begin{figure}[t]
\centering
\includegraphics[width=1\linewidth]{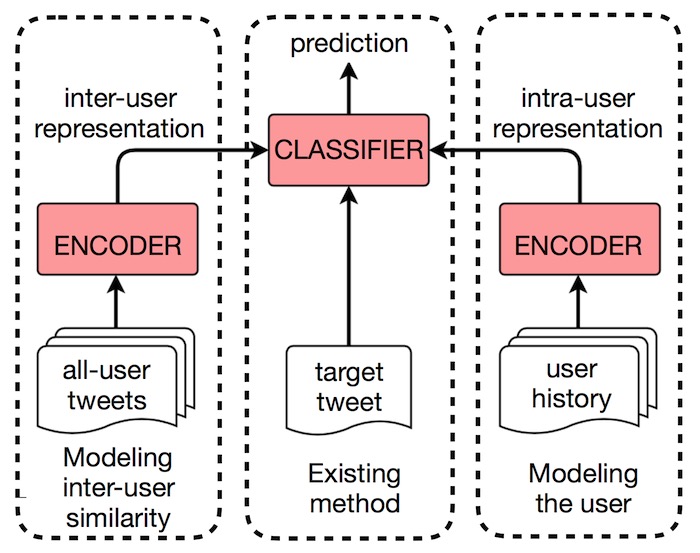} 
\caption{Our hate speech classifier. In contrast to existing methods that focus on a single target Tweet as input (center), we incorporate intra-user (right) and inter-user (left) representations to enhance  performance.
%william{Make it more compact.}
}
\label{fig:contribution}
\end{figure}

An extensive body of work has focused on the development of automatic hate speech classifiers. A recent survey outlined eight categories of features used in hate speech detection~\cite{schmidt2017survey}: 
simple surface~\cite{
%chen2012detecting,xu2012learning,warner2012detecting,sood2012automatic,burnap2015cyber,van2015detection,burnap2016us,hosseinmardi2015detection,nobata2016abusive
warner2012detecting,waseem2016hateful}, word generalization~\cite{
%xiang2012detecting,djuric2015hate,
warner2012detecting,zhong2016content}, sentiment analysis~\cite{
%dinakar2012common,sood2012automatic,gitari2015lexicon,
van2015detection}, lexical resources %~\cite{xiang2012detecting,burnap2015cyber,nobata2016abusive,spertus1997smokey,razavi2010offensive,gitari2015lexicon, burnap2016us}, 
and linguistic features~\cite{
%xu2012learning,chen2012detecting,gitari2015lexicon,burnap2015cyber,burnap2016us,zhong2016content,spertus1997smokey, nobata2016abusive
burnap2016us}, knowledge-based features~\cite{dinakar2012common}, meta-information~\cite{
%xiang2012detecting,dadvar2012improved,hosseinmardi2015detection,
waseem2016hateful}, and multi-modal information~\cite{
%hosseinmardi2015detection,
zhong2016content}. Closely related to our work is  research that leverages user attributes in the classification process such as  history of participation in hate speech and usage of profanity~\cite{xiang2012detecting,dadvar2013improving}% and gender~\cite{dadvar2012improved}
. Both~\citet{xiang2012detecting} and~\citet{dadvar2013improving} collect user history to enhance 
%\elizabeth{again, I don't think "performance" is right here.  How about "detection accuracy"?} performance. 
detection accuracy.
The former requires the user history to be labeled instances. However, labeling user history requires significant human effort. The latter models the user with manually selected features. In contrast, our approach leverages unlabeled user history to automatically model the user.

\begin{figure*}[t]
\centering
\includegraphics[width=1\textwidth]{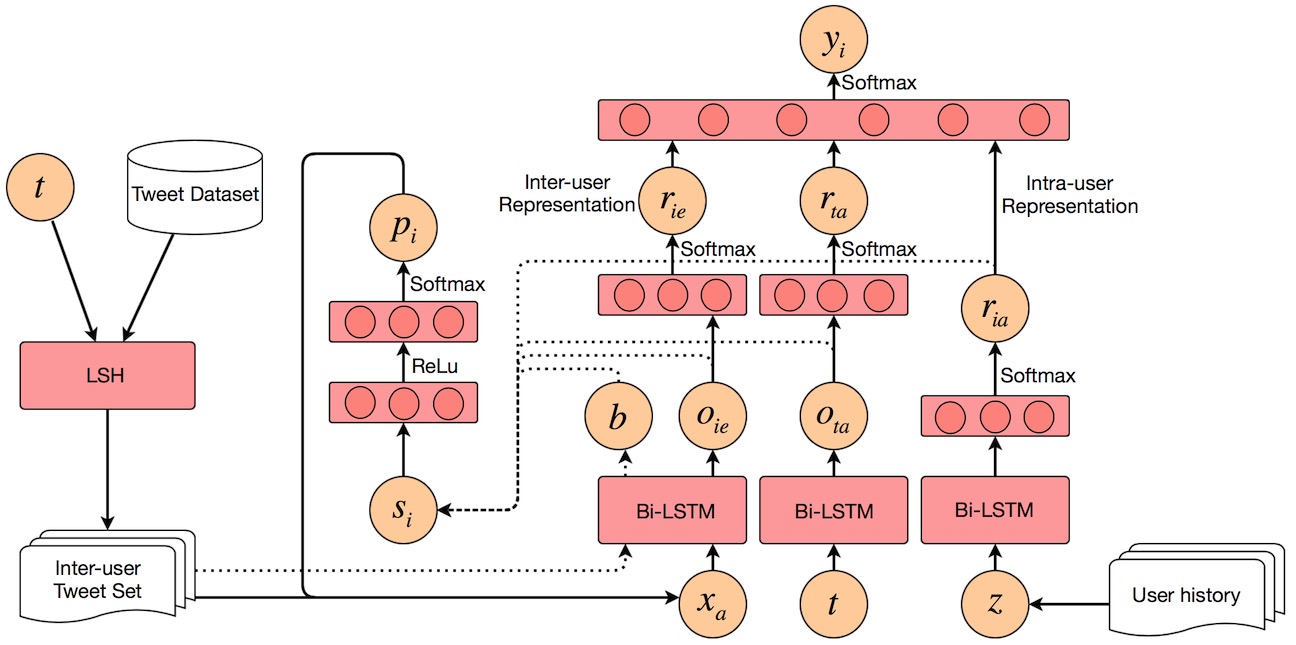}
\caption{The overview of our proposed model. $t$ is the input target Tweet, $z$ denotes intra-user Tweets, and $x_{a}$ is the selected inter-user Tweet. $r_{ie}$ is the inter-user representation, $r_{ia}$ is the intra-user representation, and $r_{ta}$ is the representation of the target Tweet. These three branches respectively correspond to the three branches illustrated in Figure~\ref{fig:contribution}. $y_i$ is the prediction at the time step $i$ and $s_i$ is the state input for the agent at the time step $i$. The  computing process is detailed in Section~\ref{sec:InterUserRepresentation}
%\william{I rememberd that you said that $r_{ie}$ and $r_{ta}$ are vectors? But in this new figure, you showed that they came from softmax and merged into a single scalar value?}
}
\label{fig:model}
\end{figure*}

In this paper, we focus on augmenting hate speech classification models by first performing representation learning to model user history without supervision.
%binary classification, which means we do not differentiate different types of hate speech. In the real world scenario, some Twitter users keep posting hate speech. Therefore, we assume that if the target tweet is hateful, then it is highly possible that there is a bunch of hateful tweets in this user's history, typically not far away from the target tweet. 
The hypothesis is that, by analyzing a corpus of the user's past Tweets, our system will better understand the language and behavior of the user, leading to better hate speech detection accuracy.
%which provides more evidence for making the final prediction. 
%\elizabeth{I think it would be worthwhile to explain what is meant by "a single Tweet is often noisy" in this context.  I could see someone not familiar with this topic thinking a tweet is either hate speech or not, and not understanding the nuances of detection.}
%\william{Please show a short example here.}
Another issue is that using a single Tweet as input is often noisy for any machine learning classifier. For example, the Tweet \textit{``I'm not sexist but I can not stand women commentators''}
%\william{Use ``'' for the correct quotation marks in latex.} 
 is actually an instance of hate speech, even though the first half is misleading. To minimize noise, we also consider semantically similar Tweets posted by other users. To do so, we propose a reinforced bidirectional long short-term memory network (LSTM) ~\cite{hochreiter1997long} to interactively leverage the similar Tweets from a large Twitter dataset to enhance the performance of the hate speech classifier. An overview of our approach is shown in Figure~\ref{fig:contribution}. The main contributions of our work are:
\vspace*{-0.05in}
\begin{itemize}
\item We provide a novel perspective on hate speech detection by modeling intra-user Tweet representations.
\vspace*{-0.05in}
\item To improve robustness, we leverage similar Tweets from a large unlabeled corpus with reinforced inter-user representations.
\vspace*{-0.05in}
\item We integrate target Tweets, intra-user and inter-user representations in a unified framework, outperforming strong baselines.
\end{itemize}
%To save space, I took this out. WW
%We introduce the approach in the next section. The dataset and experimental results are shown in Section 3. We conclude in Section 4.

\section{Approach}
%In this section, we introduce a novel approach for hate speech classification by modeling intra-user and inter-user representations. 
Figure~\ref{fig:model} illustrates the architecture of our model. It includes three branches, whose details will be described in the following subsections.

\subsection{Bidirectional LSTM}
Given a target Tweet, the baseline approach is to feed the embeddings of the Tweet into a bidirectional LSTM network~\cite{hochreiter1997long, zhou2016attention, liu2016learning} to obtain the prediction. This is shown in the middle branch in Figure~\ref{fig:contribution}.  
%\elizabeth{If you have room, can you add why this alone is not sufficient?  You start to address the limitations in the next subsection, but don't indicate why it is not sufficient}
However, this method is likely to fail when the target tweet is noisy or the critical words for making predictions are out of vocabulary. 
%\william{One thing you need to say is the out of vocabulary issue. Within a single Tweet, if there are OOV words, then the classifier will fail. }

\subsection{Intra-User Representation}
The baseline approach does not fully utilize available information, such as the user's historical Tweets. In our approach, we collect the user's historical posts through the Twitter API. For a target Tweet $t$, suppose we collect $m$ Tweets posted by this user: $Z_t=\{{z}_1, {z}_2, ..., {z}_m\}$. These intra-user Tweets are fed into a pre-trained model to obtain an intra-user representation. The pre-trained model has the same structure as the baseline model. This is shown in the right branch in Figures~\ref{fig:contribution} and~\ref{fig:model}. The intra-user representation is then combined with the baseline branch for the final prediction. The computation process is:
%shown by the following equations. 
\begin{align}
o_{ta}(t) & = f_{ta}(t,\mathbf{0})\\
r_{ta}(t) & = l_{ta}(\sigma(o_{ta}(t)))\\
o_{ia}({z}_j) & = f_{ia}({z}_j,\mathbf{0})\\
r_{ia}(t) & = \sigma({\sum_{j=1}^m}l_{ia}(\sigma(o_{ia}({z}_j)))) 
\end{align}
where $f_{ta}$ is the bi-LSTM of the baseline branch; $o_{ta}$ is the output of the bi-LSTM; and $l_{ta}$, $l_{ia}$ are linear functions. Similarly, $f_{ia}$ is the bi-LSTM of the intra-user branch and $o_{ia}$ is the output. $r_{ta}$ is the output prediction of the baseline branch. $r_{ia}$ is the intra-user representation, and $\sigma$ is the non-linear activation function. 
%\elizabeth{I suppose there is no room to explain these equations. It would be nice if we could, but perhaps that is unrealistic.}

\subsection{Inter-User Representation}
 \label{sec:InterUserRepresentation}
In addition to the user history, the Tweets that are semantically similar to the target Tweet can also be utilized to suppress noise in the target Tweet. We collect similar Tweets from large unlabeled Tweet set $U$ by Locality Sensitive Hashing (LSH) ~\cite{indyk1998approximate,gionis1999similarity}.
%\william{Jing, explain why LSH? Because the space of all Tweets is enormous...} 
Since the space of all Tweets is enormous, we use LSH to efficiently reduce the search space. 
For each target Tweet $t$, we use LSH to collect $n$ nearest neighbors of $t$ in $U$: ${x}_1$, ${x}_2$, ..., ${x}_n$. These $n$ Tweets form the inter-user Tweet set for $t$: $X_t=\{{x}_1, {x}_2, ..., {x}_n\}$. 

\begin{algorithm}[t]
\small
  \caption{Training Algorithm}\label{alg:alg1}
  \begin{algorithmic}[1]
    \For{$t$ in training set}
      \State collect $X_t$ and $Z_t$;
      \State compute intra-user representation $r_{ia}(t)$;
    \EndFor
    \State initialize parameters $\theta_p$ of the policy network;
    \State initialize parameters $\theta_e$ of the other nets;
    \For{$epoch=1, E$}
      \For{$t$ in training set}
        \State compute $o_{ta}(t)$, $r_{ta}(t)$;
        \State compute the raw prediction $y^\prime(t)$;
        \State compute $b(X_t)$;
        \State $x_a=t$;
        \State compute $o_{ie}(t)$;
        \State initialize the state $s(t)_0$;
        \For{$i=1, T$}
          \State agent select action by $\epsilon-greedy$;
          \State update $x_a$;
          \State compute $o_{ie}(t)$, $r_{ie}(t)$;
          \State compute $y(t)_i$ and $s(t)_i$;
          \State compute the reward $v_i(t)$;
        \EndFor
        \State apply REINFORCE to update $\theta_p$; 
        \State update $\theta_e$ on the loss $\mathcal{L}(\theta_e)=e(y(t)_T,y^*)$;
      \EndFor
    \EndFor
  \end{algorithmic}
\end{algorithm}

Due to the size of this set, a policy gradient-based deep reinforcement learning agent is trained to interactively fetch inter-user Tweets from $X_t$. The policy network consists of two layers as shown in the middle part of Figure~\ref{fig:model} and the policy network is trained by the REINFORCE algorithm ~\cite{williams1992simple}. At each time step $i$, the action of the agent is to select one Tweet ${x}_a$ from $X_t$.  ${x}_a$ is then fed into a bi-LSTM followed by a linear layer. The result is combined with the intra-user representation and the baseline prediction (the right and the middle branch in Figures~\ref{fig:contribution} and~\ref{fig:model}) to get the prediction at time step $i$. At each time step, the bi-LSTM layer that encodes the selected inter-user is initialized with the output hidden state of the last time step. %The process is shown in the following equations. 
The number of time steps for each target Tweet is set to be a fixed number $T$ so that the agent will terminate after $T$ fetches. The final prediction occurs at the last time step. The computation is shown by the following equations.
\begin{align}
o_{ie}(x_a)_i,&h_{ie}(x_a)_i  = f_{ie}(x_a,h_{ie}(x_b)_{i-1})\\
r_{ie}(x_a)_i  &= l_{ie}(\sigma(o_{ie}(x_a)_i))\\
y^\prime(t)  &= \sigma(l_{c}(r_{ta}(t)\oplus r_{ia}(t)))\\
y(t)_i  &= \sigma(l_{c}(r_{ie}(t)\oplus r_{ta}(t)\oplus r_{ia}(t)))
\end{align}
where $x_a$ is the selected inter-user Tweet at time step $i-1$. $f_{ie}$ is the bi-LSTM of the inter-user branch. $o_{ie}$ and $h_{ie}$ are the output and the hidden state. $l_c$ is a linear function. $r_{ie}$ is the inter-user representation. $y^\prime$ is the prediction made without the inter-user branch and $y$ is the prediction made with the inter-user branch. The symbol $\oplus$ means concatenation. The subscript $i$ denotes time step $i$.
%\william{What does vertical bar mean? If it's concatenation, then $\oplus$ is more commonly used in deep learning. Also, why $l_{c}$? I thought it should be softmax? Otherwise there's no non-linearity?}

The state at each time step for the agent is the concatenation of encoded inter-user Tweets, the output of the Bi-LSTM in the inter-user branch and the baseline branch, together with the intra-user representation in the intra-user branch (the dotted arrows in Figure~\ref{fig:model}). Each inter-user Tweet ${x}_j$ in $X_t$ is encoded by the bi-LSTM of the inter-user branch (the dotted arrow through the Bi-LSTM of the inter-user branch).
\begin{align}
& b({x}_j) = f_{ie}(x_j,\mathbf{0})\\
& s(t)_i[j]\! =\! o_{ie}(x_a)_i\oplus b({x}_j)\!\oplus o_{ta}(t)\!\oplus \!r_{ia}(t)
\end{align}
$b$ is the output of the bi-LSTM of the inter-user branch. In order to differentiate with $o_{ie}$ mentioned above, we use $b$. $s(t)_i[j]$ is the $j$th row of the state at time step $i$.

By using reinforcement learning, the state for the agent is updated after each fetch of the inter-user Tweet. Thus, the agent can interactively make selections and update the inter-user representations step by step.
The reward $v_i$ for the agent at time step $i$ is based on the original prediction without the agent and the prediction at the last time step with the agent. The computation is shown as:
%the following equations.
\begin{align}
q(t)_i & = e(y^\prime(t),y^*)-e(y(t)_T,y^*)\\
v(t)_i & =
\begin{cases}
\alpha*q(t)_i &\mbox{if $y^\prime(t)!=y^*$}\\
q(t)_i &\mbox{else if $y(t)_T!=y^*$}\\
0 &\mbox{otherwise}
\end{cases}
\end{align}
where $e$ is the loss function; $q(t)$ is the basic reward; and $v(t)_i$ is the modified reward at time step $i$. 
$\alpha$ is a positive number used to amplify the reward when the original classification is incorrect. The intuition of this reward is to train the agent to be able to correct the misclassified Tweets. When the original prediction and the last prediction are both correct, the reward is set to 0 to make the agent focus on the misclassified instances.

The complete training process is shown in Algorithm~\ref{alg:alg1}. Before the training, the intra-user Tweets and inter-user Tweets are collected for each target Tweet. Then intra-user representations are computed, followed by the computation for initializing the environment and state for the agent. Next, the agent's actions, state updates, prediction, and reward are computed. Finally, the parameters are updated.

\section{Experiments}
%In this section, we first describe the experimental setup and then  show the results of different settings.
\begin{table}[t!]
\centering
\small
\begin{tabular}{|p{4.18cm}|c|c|c|}
  \hline
   %%& \multicolumn{3}{|c|}{overall}& \multicolumn{3}{|c|}{hate speech}   \\
 %%  \cline{2-7} 
  Method & Prec. & Rec. & F1  \\
  \hline
  %LSTM &0.6721 &0.7365 &0.7029 \\
  SVM &\textbf{.793} &.656 &.718 \\
  \hline
  Logistic Regression &.782 &.611 &.686 \\
  \hline
  Bi-LSTM + attention &.760 &.665 &.710 \\
  \hline
  CNN-static &.701 &.707 &.703 \\
  \hline
  CNN-non-static &.743 &.699 &.720 \\
  \hline
  N-gram &.729 &\textbf{.777} &.739\\
  \hline
  Bi-LSTM &.672 &.737 &.703 \\
  \hline
  
  %+intra   &0.7716 &0.7485 &0.7599 \\
  + Intra. Rep.   &.772 &.749 &.760 \\
  \hline
  + Intra.+ Randomized Inter. Rep. &.773 &.764 &.768\\
  %  +intra+randomized inter &0.7727 &0.7635 &0.7681\\
  \hline
  + Intra.+ Reinforced Inter. Rep. &.775 &.773 &\textbf{.774}\\
  %  +intra+reinforced inter &0.7748 &0.7725 &0.7736\\
  \hline
\end{tabular}
\caption{Experimental results. Prec.: precision. Rec.: recall. F1: F measure. Bi-LSTM: the baseline bidirectional LSTM model. Bi-LSTM + attention: an attentional bidirectional LSTM model. The experimental settings of the last three rows are illustrated in Section~\ref{sec:ExperimentalSettings}. + Intra. Rep.: the model consists of the target Tweet branch and the intra-user branch. + Intra. + Randomized Inter. Rep.  incorporates randomly selected inter-user Tweets while + Intra. + Reinforced Inter. Rep.  further incorporates the reinforced inter-user branch. The best results are in bold.}
%\william{Missing: 1) statistical significance test 2) compare to Yoon Kim's CNN model 3) Compare with related recent work}%\elizabeth{what is F1?}}
%Jing: the statistical significance test result is stated in section 3.2
\label{tab:results}

\end{table}
\subsection{Experimental Settings}
\label{sec:ExperimentalSettings}
%\subsubsection{Dataset}
\noindent
{\bf Dataset:}
We use the dataset published by \citet{waseem2016hateful}. This dataset contains 16,907 Tweets. The original dataset only contains the Tweet ID and the label for each Tweet. We expand the dataset with user ID and Tweet text. After deleting the Tweets that are no longer accessible, the dataset we use contains 15,781 Tweets from 1,808 users.
%~\william{Explain why you need to remove part of the dataset. Maybe move the last few sentences from this paragraph to here for the explanation.} 
The published dataset has three labels: racism, sexism and none. Since we consider a binary classification setting, we union the first two sets. In the final dataset, 67\% are labeled as non-hate speech, and 33\% are labeled as hate speech. 1000 Tweets are randomly selected for testing and the remaining 14,781 Tweets are for training.
%\william{Let's try to have three subsections to save some space. 1. Baseline 2. Leveraging Intra-User representation learning. 3. Combining with Inter-User representation.}\elizabeth{I agree.  A subsection should never be just a single paragraph.}

%\subsubsection{Baseline}
\noindent
{\bf Baseline:}
The baseline model is a bi-LSTM model. The input for the model is the word embeddings of the target Tweet. The word embedding is of size 200. The hidden size of the LSTM is 64. The optimizer is Adam and we use mini-batches of size 25. The word embedding model is pre-trained on a Tweet corpus containing 3,433,513 Tweets.
%william{Are you using pre-trained word embeddings or randomly initialized?}

%\subsubsection{Leveraging Intra-user Representation Learning}
\noindent
{\bf Intra-user Representation Learning:}
Based on the target Tweet, we collect at most 400 Tweets posted by the same user, with the target Tweet removed.
%200 Tweets posted by the same user 
%before the target Tweet and at most 200 Tweets  by the same user after the target Tweet. Thus, the user history for each target Tweet consists of at most 400 Tweets. 
%The pre-trained model to encode this user history is the baseline model with best performance. 
The baseline branch and the intra-user branch are combined via a linear layer. 

\noindent
%\subsubsection{Combining with Inter-user Representation}
{\bf Combining with Inter-user Representation:}
The inter-user Tweet set is collected from the dataset via Locality Sensitive Hashing (LSH). 
%To quicken the training process, we only collect Tweets from the training dataset, but actually all the Tweets available (whether labeled or unlabeled) can be used. 
In our experiments, we use a set size of either 50, 100 or 200 Tweets. At each time step, one Tweet is selected from the inter-user Tweet set by the policy agent. %\elizabeth{what is a policy net?  Do you mean network?} 
%WW: no need to repeat the method here in the experiment section.
%The network to encode the selected Tweet has the same structure and same parameters as the baseline branch. Finally the three branches are combined by a linear layer to obtain the prediction. 
%\elizabeth{can you give your model a name so we can refer to it more easily?} 
We also experimented with a second setting, in which we replace the agent by random selection. At each time step, an inter-user Tweet is randomly selected from $X$ and fed into the inter-user branch.

\subsection{Results}
We compare the above settings with six classification models: Support Vector Machine (SVM)~\cite{suykens1999least}, Logistic Regression, attentional BI-LSTM, two CNN models by~\citet{kim2014convolutional}, and a N-gram model~\cite{waseem2016hateful}. We evaluate these models on three metrics: precision, recall and F1 score. The results are shown in Table~\ref{tab:results}.
We report results for $|U|=100$ in Table~\ref{tab:results}, as results with sizes 50 and 200 are similar.
We find that leveraging the intra-user information helps reduce false positives.
The performance is further improved when integrating our model with inter-user similarity learning. Our results show that selection by the policy gradient agent is slightly better than random selection, and we hypothesize the effect would be more salient when working with a larger unlabeled dataset. The McNemar's test shows that our model gives significantly better (at $p<0.01$) predictions than the baseline bi-LSTM and attentional bi-LSTM.   

\subsection{Error Analysis}
%To further understand our approach, an error analysis is conducted on the testing results. 
There are two types of hate speech that are misclassified. The first type contains rare words and abbreviations, e.g. \textit{FK YOU KAT AND ANDRE! \#mkr}. Such intentional misspellings or abbreviations are highly varied, making it difficult for the model to learn the correct meaning.  The second type of hate speech is satire or metaphor, e.g. \textit{Congratulations Kat. Reckon you may have the whole viewer population against you now \#mkr}. Satire and metaphors are extremely difficult to recognize. In the above two cases, both the baseline branch and the inter-user branch can be unreliable.

\section{Conclusion}
%\elizabeth{A conclusion should not be a restatement of what was already said.  You should use it as an opportunity to shed additional insight on the implications of your results, or the way in which the results could be used.  I suggest you delete this and try again.} 
In this work, we propose a novel method for hate speech detection. We use bi-LSTM as the baseline method. However, our framework can easily augment other baseline methods by incorporating intra-user and reinforced inter-user representations. 
In addition to detecting potential hate speech, our method can be  applied to help detect suspicious social media accounts. Considering the relationship between online hate speech and real-life hate actions, our solution has the potential to help analyze real-life extremists and hate groups.
Furthermore, intra-user and inter-user representation learning can be generalized to other text classification tasks, where either user history or a large collection of unlabeled data are available.
\section*{Acknowledgments}
The first author is supported by the Chancellor's Fellowship. We gratefully acknowledge the support of NVIDIA Corporation with the donation of one Titan X Pascal GPU used for this research.
\bibliography{naaclhlt2018}
\bibliographystyle{acl_natbib}

\appendix
\end{document}